\documentclass[11pt]{article}

\usepackage{acl}

\usepackage{times}
\usepackage{latexsym}
\usepackage[T1]{fontenc}

\usepackage[utf8]{inputenc}

\usepackage{microtype}

\usepackage{inconsolata}

\usepackage{booktabs}     
\usepackage{graphicx}     
\usepackage{lscape}       
\usepackage{graphicx}    
\usepackage{subcaption}  
\usepackage{caption}
\usepackage{arydshln}
\usepackage{multirow}
\usepackage{subcaption}
\usepackage{rotating} 
\usepackage{tcolorbox}
\tcbuselibrary{breakable}
\usepackage{enumitem}
\usepackage{xcolor}
\usepackage{float}
\definecolor{coldpink}{RGB}{245,190,230}

\usepackage{adjustbox}  
\usepackage{arydshln}
\usepackage{placeins} 
\usepackage{listings}
\usepackage[table]{xcolor}
\tcbuselibrary{skins, listings}

\title{Evaluating Arabic Large Language Models: A Survey of Benchmarks, Methods, and Gaps}

\author{Ahmed Alzubaidi, Shaikha Alsuwaidi, Basma El Amel Boussaha,  Leen AlQadi \\ {\bf Omar Alkaabi, Mohammed Alyafeai, Hamza Alobeidli, Hakim Hacid} \\
        Technology Innovation Institute, Abu Dhabi, UAE \\ \texttt{ahmed.alzubaidi@tii.ae}}

\begin{document}
\maketitle

\begin{abstract}
This survey provides the first systematic review of Arabic LLM benchmarks, analyzing 40+ evaluation benchmarks across NLP tasks, knowledge domains, cultural understanding, and specialized capabilities. We propose a taxonomy organizing benchmarks into four categories: Knowledge, NLP Tasks, Culture and Dialects, and Target-Specific evaluations. Our analysis reveals significant progress in benchmark diversity while identifying critical gaps: limited temporal evaluation, insufficient multi-turn dialogue assessment, and cultural misalignment in translated datasets. We examine three primary approaches: native collection, translation, and synthetic generation discussing their trade-offs regarding authenticity, scale, and cost. This work serves as a comprehensive reference for Arabic NLP researchers, providing insights into benchmark methodologies, reproducibility standards, and evaluation metrics while offering recommendations for future development.
\end{abstract}

\section{Introduction}

In recent years, Large Language Models (LLMs) have demonstrated remarkable advances in natural language understanding and logical reasoning capabilities, bringing us closer to the vision of AGI \cite{naveed2025comprehensive}. From the beginning of the transformer era, LLMs have shown potential beyond English, with numerous successes across multiple languages \cite{huang2024survey}. The Arabic language, spoken by almost 500 million people globally\footnote{World Bank (2024). Population, Total – Arab World. \url{https://data.worldbank.org/indicator/SP.POP.TOTL?locations=1A}}, exemplifies the critical need for multilingual LLM capabilities. Consequently, both bilingual and Arabic-specific LLMs have been released as closed and open-source models \cite{allandscape}. A critical component of LLM development is the availability of robust benchmarks that enable systematic evaluation of performance across tasks and domains. In parallel with progress in Arabic LLMs, a corresponding effort has emerged focused on creating comprehensive benchmarks for Arabic evaluation.

Developing Arabic benchmarks presents several unique challenges. Data scarcity and limited diversity in Arabic web content \cite{allandscape} increase the cost and effort required for benchmark creation. To address these constraints, researchers have pursued three primary approaches: translating mainstream English benchmarks to Arabic, generating synthetic data using powerful LLMs, and collecting native Arabic content. However, each approach presents trade-offs. Translation can produce culturally misaligned benchmarks that undermine evaluation robustness. Synthetic generation risks introducing biases from the generating model and may produce questions that facilitate circular evaluation. Both approaches are susceptible to cultural misalignment problems \cite{nacar2025towards}, often requiring extensive human validation and iterative refinement. 

Additionally, Arabic's linguistic diversity spanning Modern Standard Arabic (MSA) and numerous regional dialects \cite{keleg2025revisiting}, adds complexity to comprehensive evaluation. Arabic requires dedicated benchmarking analysis due to its unique morphological complexity, 20+ dialectal varieties functioning almost as distinct languages, critical cultural sensitivity requirements that undermine translation-based approaches, severe data scarcity despite 500 million speakers, and a fragmented evaluation landscape challenges absent in languages with standardized benchmarking benchmarks like English.

Several recent surveys have examined Arabic LLMs from various perspectives \cite{mashaabi2024survey,allandscape,rhel2025large}. However, none has provided an extensive, comprehensive analysis of Arabic benchmarks and evaluation approaches. This survey addresses that gap by systematically reviewing the current landscape of evaluation techniques and benchmarking datasets for Arabic LLMs. We aim to serve as the primary reference for researchers and practitioners in Arabic NLP when evaluating their models. Our contributions are threefold: (1) we propose a taxonomy organizing existing benchmarks into four major categories; (2) we provide a detailed analysis of 40+ Arabic benchmarks, offering a comprehensive resource for the Arabic NLP community; and (3) we discuss common approaches, established evaluation tools, current trends, and critical gaps in Arabic benchmark coverage. Moreover, we provide a repository\footnote{\url{https://github.com/tiiuae/Arabic-LLM-Benchmarks}} that consolidates all the benchmarks, dataset links, code repositories, and integrated frameworks to facilitate easy access for the community.

The remainder of this paper is structured as follows: Section \ref{sec:background} presents background on Arabic LLMs and related surveys. Section \ref{sec:structure} introduces our taxonomy, with Sections \ref{sec:stem} through \ref{sec:target} examining each category in detail. Section \ref{sec:discuss} analyzes the current state and identifies gaps in Arabic benchmarks. Section \ref{sec:conclusion} concludes the paper.

\section{Background}
\label{sec:background}

This section provides the necessary context for understanding the Arabic LLM benchmarking landscape. We first categorize existing Arabic LLMs by their training approaches, then review related survey work to position our contribution within the broader literature.

\subsection{Arabic LLMs}
State-of-the-art Arabic LLMs can be categorized into three types. \textbf{Native models} are trained from scratch exclusively on Arabic data, enabling Arabic-only interaction. Examples include Jais \cite{sengupta2023jais}, ArabianGPT \cite{koubaa2024arabiangptnativearabicgptbased}, and AraGPT \cite{antoun-etal-2021-aragpt2}. \textbf{Multilingual models} support multiple languages including Arabic, such as Qwen3 \cite{yang2025qwen3}, Gemma3 \cite{team2025gemma}, and Llama \cite{grattafiori2024llama}, as well as closed-source models like ChatGPT \cite{chatgpt} and Claude \cite{claude3}. \textbf{Adapted Arabic models} apply continued pretraining or supervised fine-tuning to existing multilingual models to enhance Arabic performance, exemplified by AceGPT \cite{huang2024acegptlocalizinglargelanguage}, SILMA \cite{silma_01_2024}, Fanar \cite{fanarteam2025fanararabiccentricmultimodalgenerative}, and Falcon-Arabic \cite{falcon-arabic}.

\subsection{Related Work}
Several surveys have examined Arabic LLMs from different perspectives. \citet{mashaabi2024survey} provided an in-depth discussion of pretraining and fine-tuning data for Arabic LLMs, including dialectal coverage, and listed available models with details on accessibility and reproducibility. \citet{rhel2025large} surveyed pretrained Arabic LLMs with focus on classical NLP applications and benchmarks. Most recently, \citet{allandscape} presented the historical evolution of Arabic NLP, common pretraining and fine-tuning strategies, and current research trends and challenges. While \citet{allandscape} briefly discussed a subset of existing benchmarks, no comprehensive survey of Arabic LLM benchmarks exists. This work fills that gap by systematically reviewing evaluation techniques and benchmarking datasets for Arabic LLMs.

\section{Benchmarks Taxonomy}
\label{sec:structure}
\begin{figure*}[t]
\centering
  \includegraphics[width=1.0\linewidth]{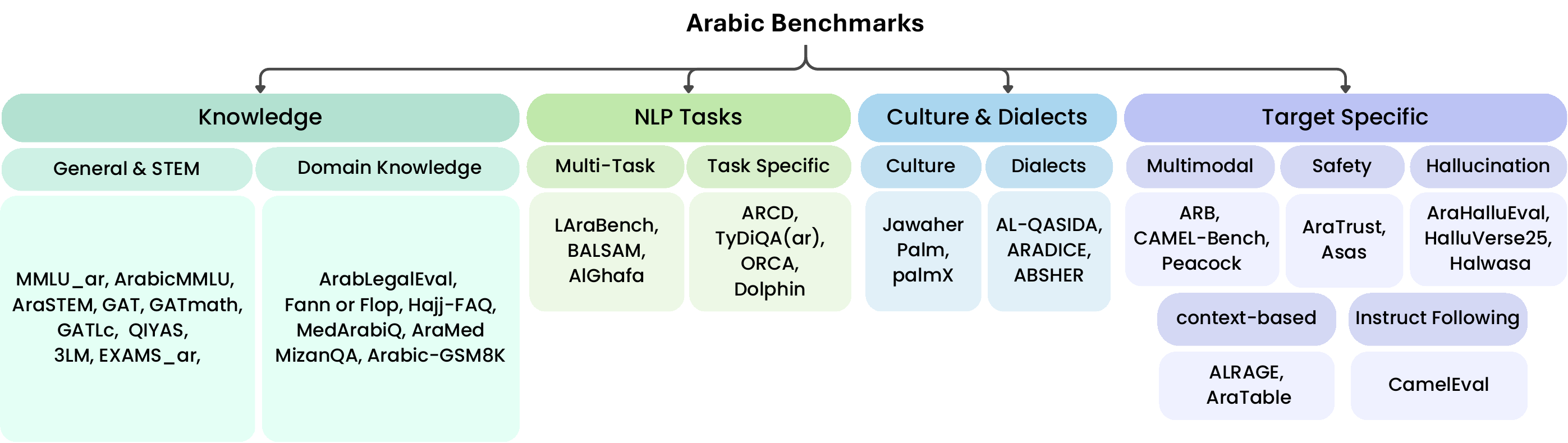} 
  \caption {Taxonomy of Arabic Benchmarks.}
  \label{fig:flow}
\end{figure*}

We reviewed 40+ existing Arabic LLM benchmarks and constructed a taxonomy that captures various themes and categories, as depicted in Figure~\ref{fig:flow}. Our taxonomy organizes benchmarks into four categories:

\noindent\textbf{Knowledge} includes benchmarks evaluating general knowledge and STEM capabilities, along with domain-specific benchmarks in fields such as law and medicine.

\noindent\textbf{Natural Language Processing (NLP)} 
encompasses early task-specific benchmarks and comprehensive multi-task benchmarks, reflecting the evolution from narrow task evaluation to unified assessment across diverse dialects and domains.

\noindent\textbf{Culture and Dialects} groups benchmarks assessing cultural knowledge and dialect understanding, addressing the essential property of cultural awareness in Arabic LLMs.

\noindent\textbf{Target-Specific} covers benchmarks designed to assess particular LLM properties such as safety, hallucination detection, instruction-following, and vision capabilities.

This taxonomy emerged from analyzing common patterns across benchmarks and reflects the evolution from task-specific evaluation to comprehensive assessment. In the following sections, we describe each category and we provide a comprehensive table (Tables \ref{tab:summary}) listing all discussed benchmarks with their characteristics.

\section{Knowledge}
\label{sec:stem}

This section examines benchmarks evaluating LLMs' acquired knowledge and reasoning capabilities, covering both general and STEM topics and specialized domains like law, medicine, and poetry.

\subsection{General and STEM}

Multilingual efforts produced early general knowledge benchmarks with Arabic components. MMLU\_ar \cite{hendrycks2020measuring} comprises 14,079 human-translated MCQs spanning 57 subjects across difficulty levels, while EXAMS\_ar \cite{hardalov2020exams} contains 562 high-school exam questions covering physics, chemistry, and biology. However, these suffer from translation concerns or limited scale.

ArabicMMLU \cite{koto2024arabicmmlu} addressed these limitations with 14,575 native Arabic MCQs curated from educational exams across Arab countries, covering all school levels plus university, spanning STEM, social sciences, humanities, and Arabic language understanding. AraSTEM \cite{mustapha2024arastem} further specialized in STEM with 11,637 native MCQs which is 7,000 more STEM samples than ArabicMMLU though it remains unpublished despite evaluation results on open-weight models.

The 3LM suite \cite{boussaha20253lm} combines three benchmarks totaling 3,151 questions: 3LM\_nat from Arabic STEM exams, 3LM\_syn synthetically generated via Yourbench's pipeline \cite{shashidhar2025yourbencheasycustomevaluation} using Qwen3-235B-A22B, and 3LM\_code comprising Arabic translations of HumanEval \cite{chen2021codex} and MBPP \cite{austin2021programsynthesislargelanguage} instructions and comments.

Several benchmarks leverage Saudi Arabia's General Aptitude Test (GAT), which assesses verbal abilities (reading comprehension, contextual errors, sentence completion) and quantitative skills (arithmetic, algebra, geometry, data analysis). Early efforts \cite{alkaoud2024bilingual,al2024qiyas} used 456 and 2,407 GAT samples respectively but lacked reproducibility and scale, evaluating only GPT-3.5 and GPT-4 with limited shots. GATmath and GATLc \cite{alballaa2025gatmath} addressed these issues with 7k and 9k samples respectively (16k total), publicly released with 5-shot evaluation on diverse Arabic and bilingual LLMs. Arabic-GSM8K \cite{arabic-gsm8k} provides human-validated translations of the established GSM8K benchmark \cite{cobbe2021gsm8k}, assessing middle-school mathematical reasoning through 5-shot settings.

\subsection{Domain Knowledge}

\textbf{Legal Domain.} ArabLegalEval \cite{hijazi2024arablegaleval} pioneered Arabic legal LLM evaluation using documents scraped from Saudi Arabia's Ministry of Justice and Board of Experts websites. The benchmark employs three approaches: synthetic MCQ generation using GPT-4 and Claude-3-opus with in-context examples from ArabicMMLU's law section, QA pairs from governmental FAQ sections, and machine-translated datasets from Legalbench \cite{guha2023legalbench} verified by legal experts. MIZANQA \cite{bahaj2025mizanqa} extends legal evaluation to Moroccan law using MCQs from various law exams.

\noindent\textbf{Poetry and Linguistics.} Fann or Flop \cite{alghallabi2025fann} assesses poetry understanding through 6,984 poem-explanation pairs, evaluating metaphorical and figurative comprehension. Evaluation uses BLEU, chrF(++), BERTScore, and mDeBERTaV3 for character-level overlap and semantic alignment, with GPT-4o as judge for faithfulness, grammatical correctness, and interpretive depth.

\noindent\textbf{Medical Domain.} \citet{al2024ahd} introduced 808k medical QA samples from the AlTibbi patient-doctor forum\footnote{\url{https://altibbi.com/}}. AraMed \cite{alasmari2024aramed} refined this to 270k high-quality samples based on vote counts. MedArabiQ \cite{daoud2025medarabiq} further curated 100 AraMed samples, enhanced them through grammatical correction and LLM modification, and added medical exam questions (MCQs and fill-in-the-blank), yielding 700 samples evaluating Arabic medical knowledge.

\noindent\textbf{Religious Domain.} Hajj-FAQ \cite{aleid2025hajj} addresses religious knowledge through question-answering on Hajj fatwas (Islamic legal rulings), providing a benchmark for LLMs' understanding of Islamic jurisprudence and pilgrimage-related guidance. This specialized dataset evaluates models' ability to handle religiously and culturally sensitive content requiring both linguistic competence and domain-specific religious knowledge.

\section{NLP Tasks}
\label{sec:nlp}

Arabic LLM evaluation on fundamental NLP tasks has evolved from narrow, task-specific datasets to comprehensive, multi-dimensional benchmarks. Early efforts like ARCD \cite{DBLP:journals/corr/abs-1906-05394} for reading comprehension and TyDiQA \cite{tydiqa} for question answering established foundational paradigms for the pre-LLM era. The landscape has shifted toward unified benchmarks assessing generalist capabilities across understanding, generation, and reasoning tasks, addressing morphological complexity, dialectal variation, and domain diversity (Table~\ref{tab:summary}).

\subsection{Comprehensive Multi-Task Benchmarks}

LAraBench \cite{larabench} pioneered systematic LLM benchmarking for Arabic NLP and speech processing, comparing task-specific models against general-purpose LLMs across 33 tasks spanning 61 datasets. It integrates earlier resources like ARCD and dialect identification corpora, bridging pre-LLM and LLM evaluation eras. Evaluation of GPT-3.5-turbo, GPT-4 \cite{openai2024gpt4technicalreport}, BLOOMZ \cite{muennighoff2022crosslingual}, Jais-13b-chat \cite{sengupta2023jais} and Whisper \cite{whisper} under zero-shot and few-shot settings reveals that specialized models outperform LLMs in zero-shot scenarios, though larger LLMs with few-shot adaptation substantially narrow this gap.

BALSAM \cite{balsam} provides community-driven comprehensive evaluation emphasizing instruction-following across 78 tasks and 14 categories, totaling 52K examples covering summarization, question answering, information extraction, translation, classification, creative writing, and reasoning. It addresses data contamination through blind test sets and provides an integrated leaderboard. Notably, LLM-as-judge correlates more strongly with human judgments (0.824 - 0.977) than traditional metrics (BLEU, ROUGE, BERTScore). Datasets combine naturally authored, translated, prompted, and synthetic sources from xP3 \cite{xp3}, PromptSource \cite{promptsource}, SuperNaturalInstructions \cite{supernaturalinstructions}, TruthfulQA \cite{truthfulqa}, and 16 newly developed datasets (1,755 prompts). Results show large closed-source models (GPT-4o, Gemini 2.0, DeepSeek V3) outperform smaller Arabic-centric models, with performance depending on tokenization quality, training data volume, and Arabic-specific fine-tuning. Limitations include potential cultural misalignment in translated datasets and insufficient assessment of multi-turn dialogue and hallucination.

\subsection{Task-Specific Benchmarks}

Understanding and generation capabilities require specialized evaluation beyond comprehensive suites.

ORCA \cite{orca} focuses on natural language understanding across Arabic's linguistic diversity, consolidating 60 datasets into 7 task clusters: sentence classification, structured prediction (NER, POS), semantic textual similarity, paraphrase identification, natural language inference, word sense disambiguation, and question answering. Evaluation employs task-appropriate metrics (accuracy, F1-score, Pearson correlation, exact match) across MSA and dialects, comparing 18 multilingual and Arabic-specific pre-trained models. Its public leaderboard with rich metadata promotes transparency and reproducibility, providing a critical reference for understanding evolution toward modern LLM evaluation.

Dolphin \cite{dolphin} addresses natural language generation through nine tasks: headline and title generation, question generation, paraphrasing, transliteration, abstractive and extractive summarization, and grammatical error correction across MSA and dialects. Consolidating 15 datasets, it evaluates encoder-decoder (mT5, AraT5) and decoder-only models (GPT-3.5, BLOOMZ, Jais) using BLEU, ROUGE, METEOR, BERTScore, and task-specific metrics like character error rate. Results reveal substantial variability and gaps between general-purpose multilingual models and Arabic-finetuned baselines. However, Dolphin predates instruction-tuned LLMs and relies on zero-shot prompting, potentially underestimating achievable performance.

Modern benchmarks like LAraBench, BALSAM, ORCA, and Dolphin provide unified evaluation benchmarks capturing understanding, generation, reasoning, and instruction-following while incorporating dialectal and domain diversity, offering a roadmap for future Arabic LLM evaluation.

\section{Culture and Dialects}
\label{sec:culture}



Evaluation of cultural and dialectal understanding in Arabic LLMs has developed progressively, with each benchmark addressing gaps left by its predecessors. Jawaher \cite{jawaher2025} provided one of the earliest culturally grounded resources by compiling 10,037 Arabic proverbs drawn from online resources. Each proverb was annotated with its dialectal origin and idiomatic explanation, and cast into a QA format, enabling evaluation of figurative and cultural reasoning in dialectal Arabic a dimension overlooked in prior benchmarks. However, Jawaher was limited to proverbial knowledge. 

PALM \cite{palm2025} addressed this gap by curating a broader culturally inclusive dataset through a year-long community-driven effort spanning all 22 Arab countries. Covering prose, dialogue, and cultural expressions across 20 topics, it expanded evaluation beyond proverbs and tested inclusivity using perplexity and generation quality. Following the release of Palm, PalmX \cite{palmx2025} released a benchmark focused on assessing the deep understanding of Arabic and Islamic culture. It contained two subtasks of MCQs in MSA covering traditions, food, history, religious practices, and expressions. Despite these advances, benchmarks had yet to capture Arab-specific commonsense reasoning. Commonsense Reasoning in Arab Culture \cite{commonsense2025} filled this gap by introducing questions requiring culturally aware inference, complementing the focus on linguistic diversity in earlier work.



We next discuss existing benchmarks that evaluate LLMs on Arabic dialects. AraDiCE \cite{aradice2025} provided a benchmark dataset that assesses the identification and generation of dialects, cognitive abilities when conversing in dialects, focusing on Egyptian, Levantine, and Gulf regions. This work involved employing machine translation techniques to translate mainstream benchmarks (i.e ArabicMMLU, TruthfilQA) to LEV/EGY dialects, to assess the desired cognitive skills. A subsequent effort by Alqasida et al. \cite{alqasida2025} proposed a comprehensive framework for evaluating LLMs’ proficiency in Arabic dialects, encompassing dimensions such as identification, comprehension, generation quality, and translation between dialects and MSA. Furthermore, NADI 2024 \cite{abdul2024nadi} was employed to evaluate dialect identification tasks, alongside several translation-based evaluations to examine dialect chat  capabilities. Absher \cite{absher2025} specialized in the Saudi dialect interpretation, focusing on vocabilary, phrases, and proverbs, sourced from the Moajam website\footnote{https://ar.mo3jam.com/}.

From the reviewed benchmarks, it is evident that only a limited number of serious efforts have been made to establish evaluation suites targeting cultural alignment and dialectal diversity in Arabic LLMs. While these initiatives mark important progress toward building linguistically and culturally grounded models, further work is needed to broaden their scope and depth. Future benchmarks should enable the community to more comprehensively assess models’ capabilities in cultural understanding and effective communication across Arabic dialects.

\begin{figure*}[t]
\centering
  \includegraphics[height=0.4\textheight, width=\textwidth]{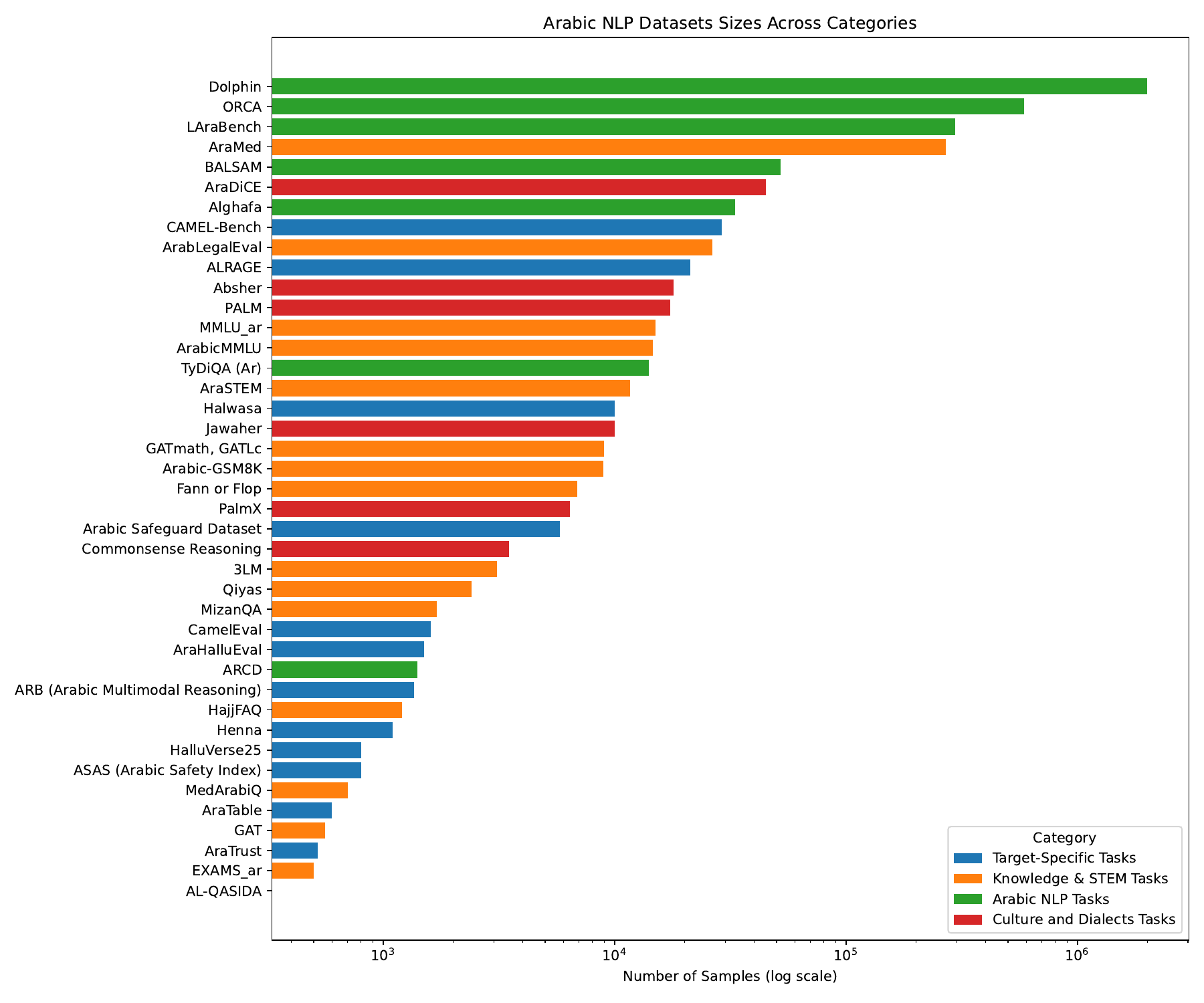} 
  \caption {Arabic Datasets Sizes Across Categories. Size is log-scaled.}
  \label{fig:category-size}
\end{figure*}

\section{Target-Specific Tasks}
\label{sec:target}

Beyond general NLP capabilities, Arabic LLMs require evaluation on specialized tasks reflecting real-world deployment scenarios and LLM-specific challenges.

\noindent\textbf{Instruction-Following.} CamelEval \cite{anonymous2025cameleval} evaluates conversational abilities and instruction-following through 1,610 generative questions: 805 human-validated translations from AlpacaEval \cite{dubois2023alpacafarm} and 805 synthetically generated using GPT-4 from culturally grounded textbooks. Evaluation uses LLM-as-judge computing win rates for open-ended generation.

\noindent\textbf{Context-Based Reasoning.} ALRAGE \cite{OALL2} targets RAG evaluation through 2.12K question-answer-context trios from 40 Arabic books, synthetically generated via Meta-Llama-3.1-70B and validated by native speakers. AraTable \cite{alshaikh2025aratablebenchmarkingllmsreasoning} addresses structured tabular data through 41 tables (15 QA pairs each) evaluating direct answering, fact verification, and complex reasoning. Tables sourced from Wikipedia, government portals, and GPT-4o were verified by human experts, with evaluation combining accuracy and the Assisted Self-Deliberation (ASD) framework employing judge LLMs and human evaluation.

\noindent\textbf{Hallucination Detection.} HalluVerse25 \cite{2025halluverse25benchmark} provides fine-grained multilingual evaluation across entity, relation, and sentence hallucination types, offering 828 Arabic samples from Wikidata autobiographies where GPT-4 injected false hallucinations validated by human judges. AraHalluEval \cite{alansari2025arahallueval} focuses on Arabic through QA (300 samples from TyDiQA-GoldP-AR and translated TruthfulQA) and summarization (100 XLSum instances), distinguishing factuality and faithfulness hallucinations through manual annotation. Halwasa \cite{mubarak-etal-2024-halwasa} provides larger-scale evaluation with 10K samples where ChatGPT and GPT-4 generated factual sentences from 1,000 SAMER Arabic Lexicon words, with hallucination indicators annotated by 200 annotators enabling sentence-level analysis.

\noindent\textbf{Safety and Trustworthiness.} \citet{ashraf2025arabicdatasetllmsafeguard} introduced 5,799 questions spanning direct attacks, indirect attacks, and harmless requests with sensitive words, employing dual-perspective evaluation from governmental and oppositional viewpoints. ASAS \cite{AIAstrolabe2025} provides the first human-rated Arabic safety index for red-teaming frontier models, exposing persistent weaknesses in top-performing systems. AraTrust \cite{alghamdi2024aratrustevaluationtrustworthinessllms} assesses trustworthiness across nine dimensions through 522 human-written MCQs.

\noindent\textbf{Multimodal Capabilities.} Peacock \cite{alwajih2024peacockfamilyarabicmultimodal} pioneered Arabic multimodal evaluation with Henna benchmark combining standard VQA/OCR prompts with focus on culture understanding. CAMEL-Bench \cite{ghaboura2024camelbenchcomprehensivearabiclmm} provides large-scale evaluation across eight domains and 38 subdomains, revealing promising performance but substantial weaknesses in specialized domains requiring precise vision-text alignment. ARB \cite{ghaboura2025arbcomprehensivearabicmultimodal} advances step-by-step multimodal reasoning, emphasizing logical integration of visual and textual inputs over simple captioning.

\section{Discussion}
\begin{table*}[t]
\centering
\resizebox{\textwidth}{!}{%
\begin{tabular}{l|l|l}
\hline
\textbf{Category} & \textbf{Coverage} & \textbf{Key Gaps} \\ 
\hline
Q\&A / Reading Comprehension & Moderate (5 datasets) & Evaluation methods vary; unified standards and human checks are needed \\ \hline
Translation \& Multitask Generation & Moderate (4 datasets) & Inconsistent task design and limited reproducibility across models \\ \hline
Reasoning \& Multi-step Thinking & \textbf{Limited} (3 datasets) & Datasets remain small and synthetic; real-world reasoning is underexplored \\ \hline
STEM / Academic Evaluation & Strong (8 datasets) & Dominated by exam-style data; lacks applied and interdisciplinary tasks \\ \hline
Law / Legal Reasoning & \textbf{Limited} (2 datasets) & Coverage restricted to few legal systems; broader legal diversity required \\ \hline
Poetry / Literature / Arts & \textbf{Limited} (3 datasets) & Creative language benchmarks remain narrow and genre-specific \\ \hline
Cultural Alignment \& Dialect Evaluation & Strong (7 datasets) & Uneven dialectal representation and limited cultural depth \\ \hline
Commonsense \& Cultural Reasoning & \textbf{Limited} (3 datasets) & Few tasks capture Arab-specific commonsense or pragmatic context \\ \hline
Hallucination / Truthfulness & \textbf{Limited} (3 datasets) & No standardized benchmark for measuring factual consistency \\ \hline
Retrieval-Augmented / Contextual Tasks & \textbf{Limited} (3 datasets) & Early research stage; grounding and factual retrieval need expansion \\ \hline
\end{tabular}%
}
\caption{Summary of Arabic NLP dataset coverage and key gaps. Coverage indicates the number of datasets per task category, while Key Gaps outline remaining needs such as standardization, scale, and cultural balance.}
\label{tab:arabic_dataset_coverage}
\end{table*}

\label{sec:discuss}

This section discusses key observations regarding reproducibility practices, community initiatives, methodological inconsistencies, and critical gaps requiring future attention.

\subsection{Reproducibility and Evaluation Benchmarks}

Reproducibility is fundamental to scientific progress, yet our survey reveals significant heterogeneity in benchmark accessibility and transparency. We classify benchmarks into three categories based on their openness:

\noindent\textbf{Private Benchmarks} release neither datasets nor evaluation pipelines, severely limiting community impact and making verification impossible. Examples include AraSTEM (11,637 samples) and Halwasa (10K samples), which remain unavailable despite published results.

\noindent\textbf{Partially-Public Benchmarks} release datasets but withhold evaluation code, hindering exact reproduction due to ambiguities in preprocessing, prompting, and metric computation.

\noindent\textbf{Public Benchmarks} release both datasets and complete pipelines, either through dedicated repositories (LAraBench, 3LM) or integration with mainstream frameworks like lighteval \cite{lighteval} and lm-eval-harness \cite{eval-harness}. 

Our analysis reveals that approximately 25\% of surveyed benchmarks remain private or partially public, limiting their impact towards Arabic NLP community. We strongly advocate for complete transparency as the community standard, recognizing that data contamination concerns can be addressed through alternative mechanisms such as blind test sets (as employed by BALSAM), periodic dataset refreshment, or held-out evaluation techniques rather than complete privatization.

\subsection{Methodological Inconsistencies and Quality Issues}

Through detailed examination of released benchmarks, we identified several concerning inconsistencies and quality issues that affect evaluation validity and cross-benchmark comparability.

\noindent\textbf{Multiple-Choice Formatting.} Benchmarks employ inconsistent option labeling: some use Latin letters while others use Arabic letters. This inconsistency affects model performance as LLMs may be more familiar with Latin alphabetic indices from English pretraining. Standardization is needed for fair cross-benchmark comparison.

\noindent\textbf{Prompting Variations.} Few-shot evaluation ranges from 0-shot to 5-shot with limited systematic investigation of optimal settings per task type. Prompt phrasing varies from formal MSA to conversational styles, potentially affecting responses.

\noindent\textbf{Quality Control Issues.} Manual inspection revealed typos, grammatical errors, and formatting inconsistencies in several released datasets. Some benchmarks contain culturally inappropriate content despite claims of cultural alignment. These quality issues undermine benchmark validity and highlight the need for rigorous review processes.

\noindent\textbf{LLM-as-Judge Validation.} While BALSAM reports 0.824-0.977 correlation with human judgments, most benchmarks adopting LLM-as-judge lack validation specifically in Arabic contexts. Judge model selection, prompt sensitivity, and potential circular evaluation when judges resemble evaluated models require systematic investigation.

\subsection{Arabic LLM Leaderboards}

Centralized leaderboards serve critical functions: establishing performance baselines, enabling fair model comparisons, tracking field progress over time, and guiding model selection for practitioners. Several initiatives have emerged to fulfill these roles for Arabic LLMs, each with distinct philosophies and design choices.

The Open Arabic LLM Leaderboard (OALL) \cite{OALL1} pioneered open-source rankings, initially using Alghafa, EXAMS\_ar, and MMLU\_ar. OALL v2 \cite{OALL2} transitioned to native benchmarks (ArabicMMLU, ALRAGE, AraTrust, MadinahQA), reflecting community consensus against translated content. BALSAM \cite{balsam} offers comprehensive evaluation across 78 tasks (52K samples) with private test sets preventing contamination, including both closed and open-source models. ILMAAM \cite{nacar2025towards} specializes in culturally aligned evaluation using refined ArabicMMLU by ensuring religious sensitivity and social norms.  The AraGen benchmark \cite{3C3H} and its associated leaderboard adopt a 3C3H evaluation metric (Correctness, Completeness, Conciseness, Helpfulness, Honesty, Harmlessness), using LLM-as-judge, and combine it with a dynamic, blind-testing approaches to push for robust and fair benchmarking of Arabic LLMs.

\subsection{Critical Gaps}

Despite substantial progress, several critical areas remain underexplored or entirely absent from current Arabic benchmarking efforts. Table \ref{tab:arabic_dataset_coverage} summarizes the coverage of datasets per category as well as key gaps.

\noindent\textbf{Underexplored Areas.} Current benchmarks lack temporal evaluation, multi-turn dialogue assessment, code-switching evaluation (Arabic-English/French mixing), low-resource dialect coverage (Sudanese, Mauritanian), pragmatic understanding (sarcasm, indirect speech), and specialized domains beyond law/medicine (education, journalism, technical documentation).

\noindent\textbf{Methodological Challenges.} Data contamination threatens validity as training corpora expand. Static benchmarks fail to capture temporal degradation. Cultural alignment lacks standardized metrics. Heavy reliance on LLM-as-judge requires more rigorous Arabic-specific validation.

\noindent\textbf{Dataset Size Matters.} As Figure \ref{fig:category-size} shows, benchmark sizes range from under 500 samples (Qiyas, EXAMS\_ar) to over 2 million (Dolphin). This dramatic variability is consequential: small benchmarks provide weaker statistical signals and are more susceptible to overfitting, while aggregate scores averaging across diverse benchmark sizes can be misleading if not properly weighted. Evaluation interpretation must account for dataset scale.

\subsection{Recommendations} 

Through this study, we would like to recommend researchers to (1) Prioritize native content with mandatory cultural review. (2) Ensure complete reproducibility through code release and framework integration. (3) Address contamination via blind test sets and adversarial filtering. (4) Standardize prompting practices and multiple-choice formatting. (5) Validate evaluation methods against diverse human judgments. (6) Expand dialectal coverage systematically. (7) Implement rigorous quality control processes. (8) Develop temporal and dynamic evaluation mechanisms.

The Arabic NLP community must coordinate on standardizing best practices, sharing evaluation infrastructure, and prioritizing underexplored areas to develop benchmarks that authentically capture Arabic's linguistic richness and cultural diversity.

\section{Conclusion}
\label{sec:conclusion}

This survey comprehensively reviews 40+ Arabic LLM benchmarks, providing the first systematic taxonomy across NLP tasks, knowledge domains, cultural understanding, and specialized capabilities. The field has evolved from task-specific datasets (ARCD, TyDiQA) to comprehensive benchmarks (BALSAM, LAraBench, ArabicMMLU) addressing understanding, generation, and reasoning.

Progress includes increased dialectal coverage, cultural grounding, and domain diversity. However, persistent challenges remain: data contamination, cultural misalignment in translations, insufficient coverage of temporal reasoning and multi-turn dialogue, and inconsistent reproducibility. Our recommendations emphasize reproducibility, cultural authenticity, dialectal inclusivity, and methodological rigor. As Arabic LLMs advance, evaluation must authentically capture Arabic's linguistic complexity and cultural richness. This survey serves as a comprehensive reference, guiding development of more robust, culturally aligned Arabic LLM evaluation.

\section*{Limitations}

Despite our efforts to conduct a comprehensive review, this survey has some limitations. The rapidly evolving nature of Arabic LLM benchmarking means recent or concurrent work may be excluded, particularly proprietary benchmarks lacking public documentation. Our review reflects benchmarks through early 2025, and given the field's accelerating pace, information may quickly become outdated. Several benchmarks mentioned in publications (AraSTEM, Halwasa, Qiyas) remain unavailable, limiting our analysis to published descriptions without independent verification. 

We focus on evaluation benchmarks rather than training datasets, emphasizing text-based and multimodal evaluations while excluding speech-only benchmarks. Our assessments of benchmark quality and cultural alignment necessarily reflect our own perspectives; native speakers from different regions may evaluate appropriateness differently. Finally, our recommendations represent informed opinions rather than community consensus, and our classification of benchmarks as public or private reflects information available at the time of writing, which may have changed subsequently.

\bibliography{acl_latex}

\newpage

\appendix

\section{Summary Table}
\label{sec:appendix}

Table \ref{tab:summary} provides a comprehensive overview of all the Arabic LLM benchmarks reviewed in this survey. This appendix clarifies abbreviations and highlights key insights for interpreting the data.

\subsection{Column Definitions}

\textbf{Sample Composition Columns:}
\begin{itemize}
    \item \textbf{NAT:} Natively authored Arabic samples (originally created in Arabic by native speakers)
    \item \textbf{SYN:} Synthetically generated samples (created using LLMs like GPT-4, Claude, or Llama)
    \item \textbf{TRAN:} Translated samples (translated from English or other languages to Arabic)
    \item \textbf{Total:} Total benchmark size
\end{itemize}

\textbf{Evaluation Details:}
\begin{itemize}
    \item \textbf{Metrics:} Evaluation metrics employed (e.g., Accuracy, F1, BLEU, ROUGE, LLM-as-Judge)
    \item \textbf{Type:} Question format - \textbf{MCQ} (Multiple-choice), \textbf{GEN} (Generative/open-ended),\textbf{BENCH} (Multi-task suite), \textbf{CL} (Classification), \textbf{RS} (Reasoning Steps)
\end{itemize}

\textbf{Accessibility:}
\begin{itemize}
    \item \textbf{PD (Public Dataset):} Yes (publicly available), No (private/request-only), Partial (subset public)
    \item \textbf{PR (Public Repository):} Yes (complete evaluation code available), No (no public repository)
\end{itemize}

\textbf{Notation:} "--" (not available/applicable), "k" (thousands), "m" (millions), "DS" (datasets)

\subsection{Key Insights from the Table}

\textbf{Dataset Composition Trends:}
\begin{itemize}
    \item \textbf{Native vs. Translated:} Early benchmarks (2020) relied heavily on translation (MMLU\_ar: 100\% translated), while recent benchmarks (2024-2025) prioritize native content (ArabicMMLU, GATmath: 100\% native), reflecting community consensus on cultural authenticity.
    \item \textbf{Synthetic Generation:} Recent benchmarks increasingly employ synthetic generation (ALRAGE: 21.2k synthetic, ArabLegalEval: 10.58k synthetic), balancing scale with cost. However, purely native benchmarks like Jawaher (10k native) and Absher (18k native) demonstrate that large-scale native collection remains feasible.
    \item \textbf{Hybrid Approaches:} BALSAM exemplifies effective hybrid strategy (26k native, 1.7k synthetic, 24k translated), combining strengths of multiple approaches.
\end{itemize}

\textbf{Scale Distribution:}
\begin{itemize}
    \item \textbf{Large-Scale:} Dolphin (2m samples) and LAraBench (296k) represent comprehensive multi-task benchmarks.
    \item \textbf{Medium-Scale:} Most benchmarks range from 1k-20k samples, balancing quality and coverage.
    \item \textbf{Specialized:} Domain-specific benchmarks (legal, medical) tend toward smaller, higher-quality datasets (1-7k samples).
\end{itemize}

\textbf{Reproducibility Concerns:}
\begin{itemize}
    \item \textbf{Fully Accessible:} Only 21 of 41 benchmarks provide both public datasets and repositories.
    \item \textbf{Dataset-Only:} 8 benchmarks release datasets without evaluation code, hindering exact reproduction.
    \item \textbf{Private:} 9 benchmarks remain completely private such as AraST EM, Halwasa, CamelEval and Qiyas, including some with substantial contributions (AraST EM: 11.6k samples), severely limiting community impact.
    \item \textbf{Critical Gap:} The high proportion of inaccessible benchmarks impedes scientific progress and independent validation.
\end{itemize}

\textbf{Evaluation Methodology Evolution:}
\begin{itemize}
    \item \textbf{Traditional Metrics:} Early benchmarks use standard metrics (Accuracy, F1, BLEU, ROUGE).
    \item \textbf{LLM-as-Judge:} Recent benchmarks increasingly adopt LLM-as-judge (BALSAM, ALRAGE, ArabLegalEval), particularly for generative tasks, though validation against human judgments remains limited.
    \item \textbf{Human Evaluation:} Hallucination benchmarks (AraHalluEval, Halwasa) maintain human-as-judge given the critical nature of the task, despite higher costs.
\end{itemize}

\textbf{Task Coverage Patterns:}
\begin{itemize}
    \item \textbf{NLP Tasks:} Well-covered with multiple comprehensive benchmarks (ORCA, Dolphin, LAraBench, BALSAM).
    \item \textbf{STEM:} Substantial progress with native benchmarks (ArabicMMLU, AraST EM, GATmath) addressing earlier translation limitations.
    \item \textbf{Culture \& Dialects:} Growing emphasis (7 benchmarks in 2025 alone), reflecting recognition of cultural alignment importance.
    \item \textbf{Target-Specific:} Emerging area with recent focus on hallucination detection and context-based reasoning.
\end{itemize}

\textbf{Temporal Distribution:}
Most benchmarks were released in 2024-2025, indicating rapid field acceleration. However, this concentration also suggests potential redundancy in some areas (multiple GAT-based benchmarks) while other areas remain unexplored (temporal evaluation, code-switching, multi-turn dialogue).

\textbf{Geographic and Cultural Bias:}
Several benchmarks focus on specific regional variants (Absher: Saudi dialect, MizanQA: Moroccan law, GATmath: Saudi standardized tests), highlighting both progress in regional representation and gaps in coverage of other Arabic-speaking regions (North Africa, Levant, Iraq).

\section{Arabic NLP Benchmarks Timeline}
\label{sec:appendix-timeline}

\begin{figure*}[t]
\centering
  \includegraphics[width=\textwidth]{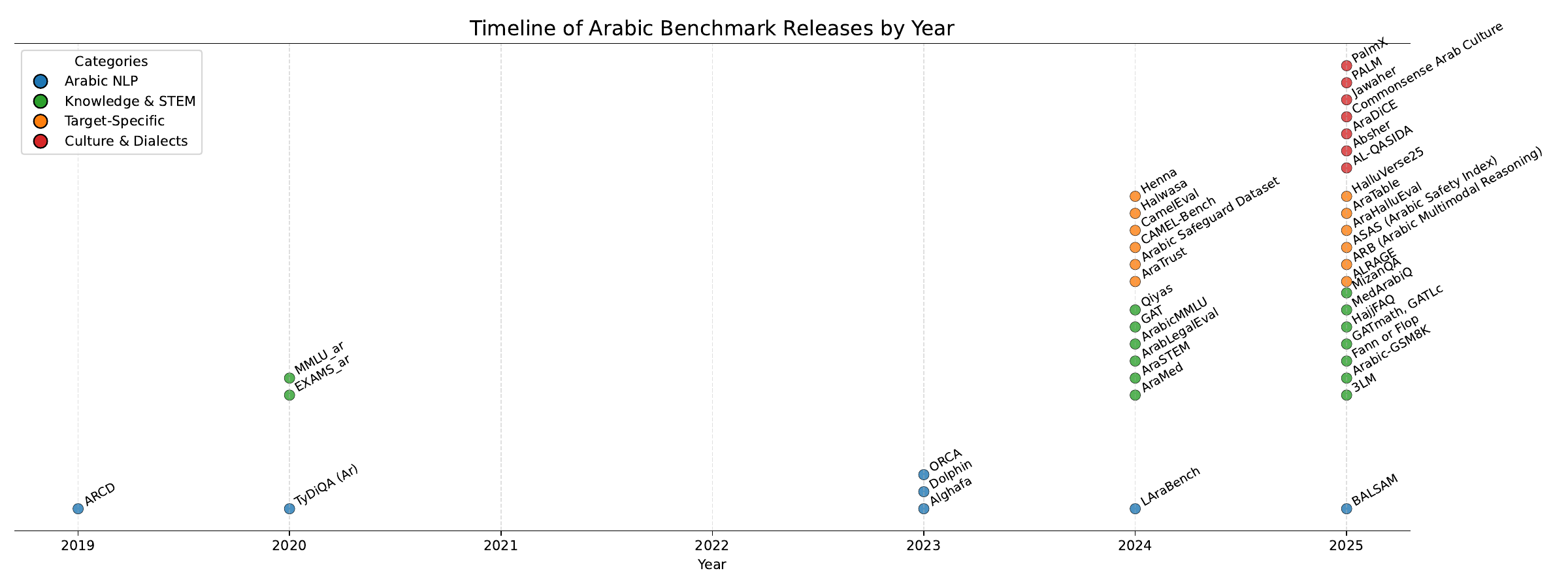} 
  \caption {Timeline of Arabic benchmark releases (2019-2025) showing dramatic acceleration, with 82\% of the benchmarks released in 2024-2025.}
  \label{fig:arabic-timeline}
\end{figure*}

Figure \ref{fig:arabic-timeline} presents a timeline of Arabic benchmark releases from 2019 to 2025, categorized by our four-category taxonomy. The visualization reveals dramatic acceleration in benchmark development, with 82\% of all benchmarks (34 of 41) released in 2024-2025 alone. This recent surge reflects growing recognition of Arabic LLM evaluation needs and demonstrates rapid field maturation across all categories, particularly in cultural/dialectal assessment where all 7 benchmarks emerged in 2025.

\begin{sidewaystable*}[t]
    \centering
    \resizebox{\linewidth}{!}{%
    \begin{tabular}{l|l|p{10cm}|c|c|c|c|c|c|c|c}
      \textbf{Year} & \textbf{Paper} & \textbf{Topic} & \textbf{NAT} & \textbf{SYN} & \textbf{TRAN} & \textbf{Total} & \textbf{Metrics} & \textbf{Type} & \textbf{PD} & \textbf{PR} \\
      \hline
            \rowcolor{gray!20}
      
      \multicolumn{11}{c}{Knowledge}\\
      \hline
      2020 & MMLU\_ar \cite{hendrycks2020measuring} & Humanities • Social Science • STEM & - & - & 15k & 15k & Accuracy • Log-prob & MCQ & Yes & Yes \\ \hline
2020 & EXAMS\_ar \cite{hardalov2020exams} & Science • Social Science • Religion & 0.5k & - & - & 0.5k & Accuracy • Log-prob & MCQ & Yes & Yes \\ \hline
2024 & ArabicMMLU \cite{koto2024arabicmmlu} & Humanities • Social Science • STEM • Arabic Language & 14.57k & - & - & 14.57k & Accuracy • Log-prob & MCQ & Yes & Yes \\ \hline
2024 & AraSTEM \cite{mustapha2024arastem} & STEM & 11.63k & - & - & 11.63k & Accuracy • Log-prob & MCQ & No & No \\ \hline
2024 & GAT \cite{alkaoud2024bilingual} & Linguistic Abilities & 0.45k & - & - & 0.56k & Accuracy & MCQ & No & No \\ \hline
2024 & Qiyas \cite{al2024qiyas} & Linguistic Abilities • Mathematics & 2.4k & - & - & 2.4k & Accuracy & MCQ & No & No \\ \hline
2024 & ArabLegalEval \cite{hijazi2024arablegaleval} & Law & 79 & 10.58k & 15.8 & 26.4k & Accuracy • LLM-as-Judge & MCQ, GEN & Yes & Yes \\ \hline
2024 & AraMed \cite{alasmari2024aramed} & Medical Domain • Healthcare  & 270k & - & - & 270k & Accuracy • QA Metrics & GEN & Yes & Yes \\ \hline
2025 & MizanQA \cite{bahaj2025mizanqa} & Law & 1.7k & - & - & 1.7k & Accuracy • F1 • ECE & MCQ & Yes & No \\ \hline
2025 & Fann or Flop \cite{alghallabi2025fann} & Poetry & 6.9k & - & - & 6.9k & BLEU • chrf(++) • BERTScore & GEN & Yes & Yes \\ \hline
2025 & GATmath, GATLc \cite{alballaa2025gatmath} & Linguistic Abilities • Mathematics & 9k & - & - & 9k & Accuracy & MCQ & Yes & No \\ \hline
2025 & 3LM \cite{boussaha20253lm} & STEM • Coding & 0.8k & 1.74k & 0.54k & 3.1k & Accuracy • Log-prob & MCQ, GEN & Yes & Yes \\ \hline
2025 & Arabic-GSM8K \cite{arabic-gsm8k} & Mathematics • Reasoning & - & - & 8.9k & 8.9k & Exact-Match & GEN & Yes & No \\ \hline
2025 & MedArabiQ \cite{daoud2025medarabiq} & Medical & - & 0.7k & - & 0.7k & Accuracy • BERTScore & MCQ & Yes & Yes \\ \hline
2025 & Hajj-FAQ \cite{aleid2025hajj} & Religious Domain • Islamic Jurisprudence & 1.2k & - & - & 1.2k & Accuracy • F1 & QA & No & No \\ \hline
      \rowcolor{gray!20}
      \multicolumn{11}{c}{Arabic NLP Tasks}\\
      \hline

2019 & ARCD \cite{DBLP:journals/corr/abs-1906-05394} & Reading Comprehension & 1.4k & - & - & 1.4k & EM • F1 & MCQ & Yes & Yes \\ \hline
2020 & TyDiQA (Ar) \cite{tydiqa} & Reading Comprehension & 14k & - & - & 14k & EM • F1 & GEN & Yes & Yes \\ \hline
2023 & ORCA \cite{orca} & Sentiment • Text Classification • NER • QA • Paraphrase • NLI • Dialect ID • MT & - & - & - & 588k & Accuracy • F1 • Pearson • EM & BENCH & Yes & No \\ \hline
2023 & Dolphin \cite{dolphin} & QA • Reading Comprehension • Reasoning • Math • Commonsense • Dialogue • Summarization • Paraphrasing • Writing  & - & - & - & 2m & BLEU • ROUGE • METEOR • BERTScore & BENCH & No & No \\ \hline
2023 & Alghafa \cite{almazrouei-etal-2023-alghafa} & Reading Comprehension • Sentiment Analysis • QA & 33.2k & - & - & 33.2k & Accuracy • Log-prob & MCQ & Yes & Yes \\ \hline
2024 & LAraBench \cite{larabench} & Sentiment • Topic Classification • NER • QA • Paraphrase • NLI • Dialect ID • MT • Reasoning • Summarization & 296k & - & - & 296k & F1 • BLEU • Task-specific & BENCH & Yes & Yes \\ \hline
2025 & BALSAM \cite{balsam} & MT • Transliteration • DialectMT • Simplification • QRewrite • Paraphrase • Intent • GrammarCorr • GenderRewrite • TextClass • Sentiment • Sarcasm • DialectID • Command • Summarization • SubjectGen • AnswerExt • SeqTag & 26k & 1.7k & 24k & 52k & LLM-as-Judge • BLEU • ROUGE & BENCH & Partial & Yes \\ 
  \hline

    \rowcolor{gray!20}
      \multicolumn{11}{c}{Culture and dialects Tasks}\\

      \hline
2025 & Jawaher \cite{jawaher2025} & Proverbs • Figurative QA & 10k & - & - & 10k & BLEURT • BERTScore • LLM-as-Judge • Human-as-Judge & GEN & Yes & No \\ \hline
2025 & PALM \cite{palm2025} & Prose • Dialogue • Cultural Expressions & 17.4k & - & - & 17.4k & LLM-as-Judge & GEN & Yes & Yes \\ \hline
2025 & PalmX \cite{palmx2025} & Shared Task • Arabic Culture MCQs • Islamic Culture MCQs & - & 6.4k & - & 6.4k & Accuracy, Log-prob & MCQ & Yes & No \\ \hline
2025 & Commonsense Reasoning in Arab Culture \cite{commonsense2025} & Cultural Commonsense QA & 3.5k & - & - & 3.5k & Accuracy • Log-prob  & MCQ & Yes & Yes \\ \hline
2025 & AraDiCE \cite{aradice2025} & Dialect Identification/Generation/Translation • Cognitive Abilities on Dialect * Culture & 41.8k & - & 45k & 81,8k & Accuracy & BENCH & Yes & Yes \\ \hline
2025 & AL-QASIDA \cite{alqasida2025} & Dialectal Analysis & - & - & - & - & Accuracy • Dialectal Error Rate & BENCH & No & No \\ \hline
2025 & Absher \cite{absher2025} & Saudi Dialect • Saudi Culture & 18k & - & - & 18k & Accuracy • F1  & MCQ & No & No \\ \hline
            \rowcolor{gray!20}
      \multicolumn{11}{c}{Target-Specific Tasks}\\
      \hline
      
      2024 & CamelEval \cite{anonymous2025cameleval} & Instruct-following & - & 0.8k & 0.8k & 1.6k & Win-rate, LLM-as-Judge & GEN & No & No \\ \hline
2024 & Halwasa \cite{mubarak-etal-2024-halwasa}  & Hallucination & - & 10k & - & 10k & Human-as-Judge & GEN & No & No \\ \hline
2024 &  Henna \cite{alwajih2024peacockfamilyarabicmultimodal} & Multimodal & -- & 1.1k & -- & 1.1k & LLM-as-Judge & VQA, OCR, GEN & Yes & Yes \\ \hline
2024 & CAMEL-Bench \cite{ghaboura2024camelbenchcomprehensivearabiclmm} & Multimodal & 29.0k & - & - & 29.0k & Accuracy, PATS • LLM-as-Judge  & VQA, OCR, RS, GEN & Yes & Yes \\ \hline
2024 & AraTrust \cite{alghamdi2024aratrustevaluationtrustworthinessllms} & Safety & 0.52k & - & - & 0.52k & Accuracy & MCQ & Yes & Yes \\ \hline
2024 & Arabic Safeguard Dataset \cite{ashraf2025arabicdatasetllmsafeguard} & Safety & 1.0k & - & 4.8k & 5.8k & LLM-as-Judge & GEN & Yes & Yes \\ \hline
2025 & ALRAGE \cite{OALL2} & Context-based (RAG) & - & 21.2K & - & 21.2K & LLM-as-Judge & GEN & Yes & Yes \\ \hline
2025 & AraTable \cite{alshaikh2025aratablebenchmarkingllmsreasoning} & Context-based (Tabular) & - & 0.6k & - & 0.6k & Accuray • Human-as-Juge & GEN & Yes & Yes \\ \hline
2025 & AraHalluEval \cite{alansari2025arahallueval}  & Hallucination & 0.4k & - & 0.8k & 1.5k & Human-as-Juge & GEN & Yes & Yes \\ \hline
2025 & HalluVerse25 \cite{2025halluverse25benchmark}  & Hallucination & - & 0.8k & - & 0.8k & Accuracy & CL & No & No \\ \hline
2025 & ARB (Arabic Multimodal Reasoning) \cite{ghaboura2025arbcomprehensivearabicmultimodal} & Multimodal & 1.36k & - & - & 1.36k & BLEU • ROUGE • BERTScore • LaBSE • LLM-as-Judge & VQA, RS & Yes & Yes \\ \hline
2025 & ASAS (Arab-ic Safety Index) \cite{AIAstrolabe2025} & Safety & 0.8k & - & - & 0.8k & LLM-as-Judge & GEN & No & No \\ \hline

    \end{tabular}
    }
    \caption{Summary of Arabic Benchmarks. \textbf{NAT:} \# of native samples, \textbf{TRAN:} \# of translated samples,  \textbf{SYN:} \# of synthetic samples, \textbf{PD:} dataset is public, and \textbf{PR:} repo is public.}
    \label{tab:summary}
\end{sidewaystable*}

\newpage

\end{document}